**Transformers, Contextualism, and Polysemy**

**Jumbly Grindrod, University of Reading**

**Accepted at *Ergo***

**Abstract**: The transformer architecture, introduced by Vaswani et al. (2017), is at the heart of the remarkable recent progress in the development of language models, including widely-used chatbots such as Chat-GPT and Claude. In this paper, I argue that we can extract from the way the transformer architecture works a theory of the relationship between context and meaning. I call this the *transformer theory*, and I argue that it is novel with regard to two related philosophical debates: the contextualism debate regarding the extent of context-sensitivity across natural language, and the polysemy debate regarding how polysemy should be captured within an account of word meaning.

## 1. Introduction

The recent emergence of large language models (LLMs) promises to change many aspects of society, including various work sectors and aspects of education. But could it help inform philosophical and linguistic debates regarding the nature of linguistic meaning? Some are skeptical that LLMs could provide any linguistic value (Chomsky et al. 2023; Dupre 2021), while others see the beginning of a new kind of linguistic inquiry (Baroni 2022; Piantadosi 2023). Specifically, both Baroni and Piantadosi have argued that LLMs can themselves be thought of as linguistic theories that have greater predictive power than anything else we currently possess. I will not seek to assess or defend that stronger position here. Instead, I want to consider whether one kind of LLM – the transformer architecture (Vaswani et al. 2017) that is at the heart of all of the state-of-the-art language models available today – can shed light on the relationship between context and meaning. I will do this by considering whether the way that the transformer architecture processes linguistic data is suggestive of a novel position regarding two related debates: the contextualism debate and the polysemy debate.

The contextualism debate is concerned with the extent and nature of context-sensitivity across natural language. Contextualists have argued that context-sensitivity is a general feature of natural language such that any sentence can vary in its interpretation across different contexts of use. They also typically argue that this has serious ramifications for the semantics/pragmatics divide. The polysemy debate is concerned with how polysemous expressions – expressions that appear to be ambiguous between a number of closely-related senses – are stored lexically. Both debates sit within the broader project of constructing a theory of meaning for a language. Before going into greater detail regarding contextualism, polysemy, and the transformer architecture, I want to first paint a more general picture of this broad project and the role that LLMs might play as part of it. Of course, if Baroni and Piantadosi are right, then the answer is straightforward, for LLMs themselves are linguistics theories that, as Piantadosi has argued, are not surpassed. But while the conclusions in this paper are consistent with that kind of bold approach, they don't depend on it.

One recurrent issue in theorising about meaning is the extent to which the project should be viewed as a cognitive endeavour. More generally, for any given theory there are two related questions we can ask:

1) Does a theory draw on cognitive considerations as part of its evidence base?

2) Does a theory have a cognitive phenomenon as its explanandum?

It is plausible that a positive answer to 2 entails a positive answer to 1. That is, it would be hard to imagine a theoretical project where a cognitive phenomenon is theorised about but no cognitive information informs theory construction. But a negative answer to 2 does not entail a negative answer to 1, for the types of consideration that can be included within the evidence base for a theory can be broader than the domain of the explanandum. For instance, a psychological theory might draw upon evolutionary evidence, but it does not follow that the theory is about some evolutionary phenomenon.[1]

The distinction is worth making here regarding theories of meaning, as there is a difference between the evidence that a theory of meaning draws upon and the phenomenon that it is about. Some have adopted a more cognitive view of the explanandum of a meaning theory; for instance Larson and Segal (1995) and Borg (2004) are explicit that they view the job description of semantics to be to capture semantic competence as realized within the individual speaker. Others have rejected the cognitive view (Katz 1981; Katz & Postal 1991; Soames 2008), instead arguing that the linguistic object of study is an abstract system. But even if the more cognitive approach is rejected (i.e. a negative answer to 2), it does not follow that cognitive considerations regarding semantic cognition are irrelevant to a theory of meaning. In particular, such considerations can serve as a weaker form of defeasible evidence in favour of a given theory of meaning.

The particular type of evidence I have in mind here is what we can call an *implementation* of a meaning theory.[2] The idea is that a given meaning theory is implemented within a cognitive system just in case there is a correspondence between the objects, processes, and structures posited by the meaning theory and those employed within the cognitive system. To use a big-picture example, consider what Recanati (2003: 4 ff.) has previously labelled the "standard approach" to context and meaning, where truth-conditional semantics is thought to operate largely only through the use of a lexical semantics plus some rules of semantic composition. Context is only appealed to for certain expressions (indexicals, demonstratives etc.), and sentences that do not contain those expressions can be assigned a truth-conditional meaning with no appeal to context. There are then certain post-semantic pragmatic processes that generate further content, such as conversational implicatures, and these largely follow Grice's (1989) classical account. This standard picture makes appeal to a set of structures, objects, and processes and we can explore whether there is a correspondence between those and what takes place in the ordinary speaker when they comprehend and produce sentences. For instance, the view posits a lexical store of word meanings, and some procedure of semantic composition. It claims that the output of this semantic composition procedure is an assignment of truth-conditional content to a sentence, and it claims that the output of the pragmatic procedures will be further truth-conditional content. We can then ask whether there is a correspondence between these structures, objects, and procedures and those that we find within linguistic cognition.

---

[1] As a reviewer notes, following Fodor (1980), whether a given theory ends up drawing upon cognitive considerations or not is plausibly an a posteriori matter. What I say here is consistent with that. This would be simply to claim that the answer to 1 will be discovered empirically rather than determined in advance of theorising.

[2] Here I use "meaning theory" to refer to a theory of how meaningful communication is achieved, including both what is traditionally viewed as the semantic domain and the pragmatic domain. One feature of the contextualism debate in particular is that the distinction between semantics and pragmatics is up for grabs, and I don't want my conclusions here to be restricted to consideration of semantic facts only.

Even if 2 is answered negatively, the fact that a given meaning theory is implemented in human cognition would still plausibly constitute positive evidence in favour of that theory. After all, even those who think that the proper subject matter of semantics is an abstract system would plausibly expect that speakers competent in a language will have in some sense internalized that system. It is no surprise then that cognitive considerations are standardly taken to be relevant across such debates; both the contextualism debate and polysemy debate frequently make appeal to cognitive considerations, for instance in Borg's (2004) appeal to cognitive modularity as part of her case against contextualism, or in the oft-cited processing differences between polysemous and homonymous expressions (Klepousniotou & Baum 2007).

As well as asking whether a meaning theory is implemented in a human cognitive system, we can also ask whether it is implemented in an artificial system, and this is where LLMs enter the picture. That is, if we have a system that both receives and produces linguistic signals, as LLMs do, we can ask whether the manner in which they do this makes use of structures, objects, and procedures that correspond to those posited by the meaning theory. And if it is the case that the structures, objects, and procedures do not straightforwardly correspond to a particular meaning theory, then we can work backwards to reconstruct a meaning theory that would so correspond. In what follows, I will seek to do this with regard to the transformer architecture. I will call that partial theory the *transformer theory*.

In order for this implementation question to be asked regarding LLMs, I will be working with a significant assumption that is not uncontroversial. This is that LLMs do successfully interpret the meanings of the sentences that they are given as input, and that they use the interpretation of the meaning of that sentence to inform their performance in whatever task they are set, whether it is to produce new text as a chatbot or complete some other task (e.g. classification tasks, translation task, summarization task etc.). Just as a meaning theory will assign meanings to sentences, the assumption here is that LLMs do this as part of an interpretative procedure. I will provide greater detail about how transformer models do this in §2, but for now it is enough to note that if we allow this assumption, the interpretation is achieved by assigning a vector to each word in the input text – more specifically, a particular type of vector known as a *contextualized embedding*. The reason why we need this assumption is that it provides an anchor, a fixed point of correspondence between the LLM and the meaning theory with which we can then consider whether the processes, objects, structures etc. that lead to the assignment of meaning, across theory and system respectively, correspond.

The idea that an interpretation could be achieved through vector assignments is certainly controversial but not without precedent. In particular, distributional semantics is an approach to meaning common in computational linguistics that proceeds on the idea that the distributional properties of words can be used to generate vectors of this kind where such vectors can figure as word meanings. While I will outline the distributional approach in greater detail in §2, I will not seek to explicitly defend it. For further discussion of the plausibility of the distributional approach, see: [Reference anonymized] (Lenci, 2008 2018; Coelho Mollo & Millière 2023).

But while I will proceed as if this assumption is true, there is still value in asking the implementation question even if we take a more agnostic stance towards the interpretative capabilities of LLMs. Instead of being committed to the claim that LLMs do in fact interpret linguistic meanings, we can instead treat LLMs as a model (in the scientific sense) of an interpretative system. Models are often used in scientific inquiry as proxies for the actual target phenomenon. Here, it could be allowed that LLMs may not in fact interpret linguistic inputs, but that they possess many of the same abilities as cognitive systems that do interpret linguistic inputs – as evidenced by their state-of-the-art performance across a wide range of linguistic tasks, including in chatbot capabilities.

There are a number of reasons why it is worth applying the implementation question to LLMs, even if they are understood as mere models of linguistic interpretation. First, as has been noted in philosophy of science, there is significant scientific value in mere "how-possibly" explanations, particularly but not only when "how-actually" explanations are absent (Rohwer & Rice 2013; Reutlinger et al. 2018). While scientific inquiry will typically aim to explain what actually gave rise to the target phenomenon, it is sometimes the case that, particularly at an earlier stage of inquiry, there is value in seeing that a given explanation is even a possible candidate. Schelling's (1971) famous model of racial segregation is often pointed to in this regard, as a way of showing how housing segregation is possible given certain assumptions about houseowner preference. It was never supposed to be an explanation of segregation in any particular city or country, but still provides significant insight into the phenomenon. In the case of LLMs, it is evident that they possess certain linguistic abilities – the range of abilities associated with producing long strings of coherent text across a range of different formats, topics, styles etc. If the way a given LLM has achieved this is by implementing a particular meaning theory, this provides us with a how-possibly explanation regarding those abilities. This bears direct relevance to certain issues in the contextualism debate. For instance, Cappelen and Lepore (2005: 123–128) have argued that if radical contextualism were true, communication would not be possible. So if we were to find that an artificial system that implements radical contextualism was able to gain some of the abilities required for communication, then this would start to speak against their worry.

Second, and relatedly, in considering the kind of meaning theory that has been implemented in a particular LLM architecture, it may well be that the LLM implements a meaning theory quite unlike those considered previously. Indeed, this is what I will argue, that the transformer architecture processes meaning in a way that resembles both contextualist and non-contextualist theories, as well as rival views regarding polysemy. So the implementation question regarding LLMs is useful as a way of considering alternative positions in the logical space.

Third, we are currently at a relatively early stage of exploring whether LLM technology can be implemented in a more humanly realistic way. One example of this is in the so-called "baby LM" research project, which aims to train language models using a training set that broadly reflects the amount of exposure to linguistic data that a child would have. If, further down the line, we find that language models can be produced that do meet many of the same restrictions faced by human speakers, the inferential leap from the language model case to the human case becomes narrower. Given that the transformer architecture forms the basis of the current state-of-the-art, our best current guess at what any possible baby LM would look like is that it closely resembles that architecture. So inspection of that architecture now can inform future implementation inquiries.

Before embarking on this implementation question, the limitations of this paper should be made clear. First, I will not be able to give an exhaustive overview of the various forms of large language model available today. The technological landscape is extremely fast-moving, and models differ on a range of details that, while they may be philosophically interesting in their own right, have to be overlooked for current purposes. Instead, I will focus on the transformer architecture that has been at the heart of the recent leaps of progress in LLM technology and that is adopted by all state-of-the-art models today. Secondly, my consideration of whether a meaning theory has been implemented in the transformer architecture will only be partial, for there are simply too many theoretical issues to consider for the purposes of a single paper. However, this is consistent with the verdict I aim to reach, which is that the transformer architecture implements a novel meaning theory. To do that, it is enough to highlight certain points of difference between existing theories in the literature and the meaning theory that is implemented in transformers without considering every aspect of a meaning theory.

With these limitations in mind, I will proceed as follows. In §2, I will provide an overview of how transformer models work, focusing on the self-attention mechanism as the key mechanism that captures the interaction between context and meaning. In §3, I will then position the transformer theory within the contextualism debate, arguing that contrary to initial appearances, the transformer theory cannot simply be understood as an implementation of radical contextualism. Specifically, the transformer theory allows for a notion of standing meaning that is more amenable to an opponent view of radical contextualism: semantic minimalism. In §4, I will then turn to the related debate on ambiguity and lexical semantics. Following Trott & Bergen (2023), I will distinguish between the *core representation approach* and the *meaning continuity approach*, and argue again that despite initial appearances, the transformer theory cannot be straightforwardly categorized as implementing the meaning continuity position. Instead, the transformer theory is really a combination of both types of approach. I will finish by emphasizing the strengths of the transformer theory, as a call for further inquiry into a previously-neglected position.

## 2. The transformer architecture

I will begin with an informal overview of how the transformer architecture processes word meaning, so that we can then consider the meaning theory that is implemented within it. To understand how transformers work, it is useful to first turn to an approach to meaning known as *distributional semantics*. According to distributional semantics, the meaning of a word can be represented by its distribution across a suitably large corpus. Words that are (dis)similar in meaning will have (dis)similar distributions. While this claim may strike many as odd, it is worth keeping in mind that it is treated by advocates of distributional semantics as a working hypothesis: a claim to be provisionally adopted to see what fruit it bears. Although distributional semantics has an intellectual history that stretches back at least 70 years (Firth 1957; Harris 1954), it has gained traction in more recent times by being combined with a vector space methodology. That is, the distribution of each word is captured as an ordered list of real numbers that can be represented as a point in a high-dimensional space (where the number of elements in the list corresponds to the dimensionality of the space). But how could such an ordered list represent the distribution of a word? While previously, "count" approaches have been used where elements in the vector correspond to how often the word co-occurs with other words (initially, at least), a "predict" approach is now far more common, such as the widely-used Word2vec algorithm (Baroni et al. 2014; Mikolov et al. 2013). Word2vec generates vectors for each word using a small, self-supervised neural network that is trained on a language prediction task. During training, the network will take a sentence from its training data, mask a word or a set of words, and attempt to predict the words that are masked. It will then use back-propagation along with an optimization algorithm to adjust all weights in the network in order to minimize the difference between its prediction and the correct result. This process is repeated many times, possibly working many times through the training data. The result is that each word is represented as a set of weights between the initial layer and the middle layer of the network, which can be extracted and serve as the vector representation of that word's distribution. It was something of a discovery – a partial vindication of the distributional semantic approach – that such vectors could be treated as representations of word meaning. This was not vindicated so much by the model's success in the language modelling task that it was trained on, instead it was that the resultant vectors improved the state-of-the-art across a broad range of natural language processing tasks to do with meaning. The vector-based approach is now completely ubiquitous across natural language processing.

Transformer architectures are a more sophisticated version of this vector-based approach embodied by Word2vec and similar. For our purposes, we will focus on the mechanism that is arguably central to the transformer architecture: the self-attention mechanism.[3] We can see the need for such a mechanism by considering how a set of static word vectors could be used to represent a complex expression like a phrase or a sentence. For any word in a sentence or phrase, we can retrieve its Word2vec vector (or *embedding*).[4] And so a complex expression can be represented in terms of the set of vectors associated with its constituent words or by some combination of those vectors (e.g. by summing them together). But this would seem to tell us nothing about the particular way in which those words interact with one another to form a grammatical complex expression. This includes syntactic relations i.e. the particular way in which expressions interact with one another to form a grammatical whole. This also includes all the phenomena that fall under the broad umbrella of the compositionality of meaning,[5] as well as the way particular expressions vary their meanings on particular occasions of use e.g. demonstratives, indexicals, and ambiguous expressions. Word2vec embeddings by their format must remain silent on how these issues are resolved on particular occasions of use. This is obviously a serious shortcoming in building a system that attempts to track the meanings of words and sentences.

What is needed then is a way of registering for each word the wider context in which it appeared and having that affect the way that the word is represented. A common way of introducing this kind of sensitivity is to use a *recurrent neural network* (RNN). RNNs process each datapoint coupled with a hidden state that registers the previous datapoint. This mechanism allows for each datapoint to be processed in a way that is sensitive to its wider context (this sensitivity can stretch further than its immediate predecessor because that immediate predecessor was itself processed in a way sensitive to its immediate predecessor, and so on). However, this means that distance between datapoints makes a difference, and RNNs are thought to struggle with longer distance relationships between datapoints.[6] In language, many syntactic and semantic relationships can occur over indefinitely long strings of text and so RNNs will potentially struggle to capture them. Note also that there is a practical concern in terms of how much RNNs can be scaled up in size. Because for each datapoint n, n-1 must already have been processed in order to generate the hidden state, all data needs to be processed sequentially. This sets limitations on scale, where training models and processing very large datasets becomes computationally expensive.

[3] In doing so, we will be ignoring other features of the architecture, such as the distinction between encoders and decoders (or combinations of the two), and the use of a feed-forward network to introduce non-linear processing. The reason for doing this is that, as we will see, it is in the self-attention mechanism that some element of context-sensitivity is introduced. See: (Geva et al. 2021) for an attempt to interpret what the feed-forward layers in transformer models do.

[4] "Embedding" is commonly used to refer to vectors that are produced using neural networks in the way described above. For the remainder of the paper, I will use "embedding" and "vector" interchangeably.

[5] To take just one example among many, when a verb is modified by an adverb, it is not just the case that the complex expression is grammatical: the meaning of the complex expression is a result of some interaction between the semantics of the two constituent expressions.

[6] *Long short-term memory models* are a sophisticated form of recurrent neural network designed to overcome this issue. However, they still suffer from the problem (that will be discussed in the remainder of this paragraph) that they employ sequential processing.

The transformer architecture was introduced as a way of avoiding both issues (Vaswani et al. 2017). Rather than employing a hidden state mechanism, the transformer introduces an element of context-sensitivity in a way that avoids sequential processing and that will not struggle with long-range dependencies by design.[7] I will focus here particularly on the self-attention procedure as the process that most clearly introduces context-sensitivity.

A simplified way of understanding the self-attention procedure is as follows. For each word embedding, we replace it with a new embedding determined on the basis of the original embedding plus the embeddings of the surrounding words. To do this, the original word embedding is run through three distinct linear layers to produce a query embedding, key embedding, and value embedding.[8,9] For each word, the dot product (a similarity measure) is taken between its query and every other key, and these scores are then run through softmax so that they sum to 1. These are called the self-attention scores, and they are then used as part of a weighted sum by multiplying against the value embeddings to generate a new embedding. The crucial thing to note about the self-attention procedure is that it allows for a kind of cross-pollination between words, but that how exactly this cross-pollination works is determined in the training of the model. Specifically, because each self-attention head will generate three distinct embeddings for each word using three independent linear layers, the information that each self-attention head focuses on in generating the new embedding is determined in training by setting the parameters for those linear layers.

Transformers have many layers of many self-attention heads. In each layer, once the self-attention procedure is complete and a new embedding is generated, the embeddings that all self-attention heads output are then concatenated and run through another linear layer, which allows the information generated across all self-attention heads to be combined into a single new embedding. This is then passed to a feed-forward network before this process is repeated at the next layer of self-attention heads.

The high number of self-attention heads allows for specialization in what each focuses on. In theory, some might be focused on homonymy disambiguation, others on anaphora resolution, others on particular syntactic relationships etc. There is now a thriving area of research that focuses on investigating the kind of information that self-attention heads are sensitive to (Clark et al. 2019; Devlin et al. 2019; Rogers et al. 2021). There is also considerable work being done on constructing tools for visualizing the behaviour of self-attention heads, allowing the theorist to explore the behaviour of various attention heads at various levels (Vig 2019; Wang et al. 2021).

---

7 There is still the issue of the "context window" that transformers employ, which places restriction on what language models can take into account. It may, however, be too quick to claim that the memory of a language model is equal to its context window, for the same reason that a RNN is not only sensitive to the previous datapoint. Either way, language models have been introduced in recent years with impressively large context-windows: OpenAI has introduced a range of "turbo" versions of gpt 3.5 and 4, which have context windows many thousands of tokens wide.

8 A linear layer is akin to a fully-connected neural network for which there is no hidden layer, only an input layer and an output layer.

9 The linear layer typically outputs a vector of lower dimensionality. If d is the dimensionality of the original embedding and n is the number of self-attention heads in a layer, the linear layers typically output n embedding of d/n. This means that when the self-attention embeddings are concatenated, the original dimensionality is restored.

One important feature to note about the self-attention procedure is that it is a vector-to-vector procedure: it will take a set of vectors, one for each word in the textual input, and it will generate a set of new vectors. As noted earlier, the latter are sometimes called *contextualized embeddings* because they carry information about the context that the word appeared in. But generating contextualized embeddings requires that there are vectors that are invariantly assigned to each word and that serve as input to the self-attention procedure. These are sometimes called *token embeddings*.[10] These are similar in nature to static Word2vec embeddings insofar as they are learnt through training and are assigned to each word in the model's vocabulary.

A second important feature for our purposes is that there is no restriction built in on what contextual features the self-attention head focuses, or on how it changes the word embedding on the basis of that. This is determined completely in the training procedure. An important consideration in what follows will be the unconstrained nature of the processing, particularly when considering the contextualism debate, to which I now turn.

## 3. Transformers and contextualism

In discussing how the self-attention mechanism works, I have already made appeal to context and meaning in interaction insofar as I have described the general procedure by which LLMs introduce information regarding the context a word is used in. In philosophy of language, one of the central debates regarding context-sensitivity is the so-called contextualism debate. In this section, I will first provide an overview of the debate, before turning to consider where the transformer theory sits within that debate.

The best way into the contextualism debate is to consider a few of the cases that are typically wielded in support of the view:

3) The leaves are green.
4) Stokes is ready.
5) The Netherlands is flat.

Consider a plant whose leaves would be brown if not for the green paint smothered over them (Travis 1997). On some but not all occasions, an utterance of 3 to speak of that plant would be true. Consider Stokes, who needs no further preparation before heading to the party but has not yet realized how much of a responsibility it would be to be a parent. On some but not all occasions, an utterance of 4 to speak of Stokes would be true. Finally, consider the Netherlands, one of the flattest countries in the world, but obviously not as flat as a snooker table. On some but not all occasions, an utterance of 5 will be true.

For the contextualist, the context-sensitivity exhibited by sentences like 3-5 is a general feature of language. All sentences in a language will exhibit this context-sensitivity, as the "propositional contribution of an expression is not fully determined by the invariant meaning conventionally associated with the expression type but depends upon the context" (Recanati 2010: 17). So one core part of the contextualist view is a claim about the *extent* of context-sensitivity: that the contribution *any* expression makes is a context-sensitive matter.

[10] This oversimplifies somewhat: what is actually fed into the self-attention procedure is an embedding which is the sum of the token embedding, a positional embedding (that indicates where in the input text the word appeared) and, for some models, a segment embedding which indicates in models where the input is split into two or more parts, which part the word is in. For our purposes, we will not be interested in the positional embeddings or segment embeddings.

A second important part of the contextualist view is that this context-sensitivity is not linguistically mandated in every instance. Following Recanati (2003), we can distinguish between *modulation* and *saturation*. Saturation occurs when determination of an expression's semantic value requires some contextual input, as is the case with indexicals and demonstratives. Modulation, on the other hand, occurs when the semantic value of an expression is modified on a particular occasion of use, but no such modification was required according to the standing meaning of the expression. What counts as modulation is itself a controversial topic, with various phenomena including enrichment, loosening, semantic transfer, metaphor, irony, and hyperbole all suggested as members of the set.[11] However, I will not focus on that issue here, as taxonomizing the various forms of unlicenced contextual effects on what is said will be less important than the general category of modulation.

While the term "radical contextualism" is sometimes used interchangeably with "contextualism" as I have defined it here, Recanati's (2003, 2010) distinction between the two views will be useful for our purposes.[12] Allowing for modulation as a determinant of what is said by an utterance is consistent with the idea that at least some of the time, no such modulation occurs for a particular expression or even an entire sentence – perhaps some of the time the meaning of an utterance is completely inherited from the meaning of the uttered sentence. Radical contextualism is the view that this never occurs, for expression meanings are not of the right format to figure as part of truth-conditional content. Recanati (2010: 18) characterizes the view as rejecting the Fregean presupposition that "the conventions of a language associate expressions with senses". Rather, if we are to understand senses as constituents of truth-conditional meaning, then expression meanings are only one determinant of their senses. So radical contextualism is a stronger form of contextualism insofar as it adds a further claim about the nature of expression meaning.

While we are considering the radical arm of contextualism, it is important to highlight one even more extreme view that Recanati (2003: 146–151) considers: meaning eliminativism. This is the view that expressions do not have meanings. That is, there is no dedicated lexical information stored for each expression. Instead, a speaker just relies on her encyclopaedic knowledge of the world and/or her long-term memory of previous uses of an expression in order to make a call on what each expression means on every occasion of use. The particular version that Recanati considers is an earlier proposal from Hintzman (1986) according to which hearers rely only on the previous uses of an expression as stored in their long-term memory. But the view has arguably become more popular in psycholinguistics, for instance in Elman's (2004, 2009) influential model of language comprehension according to which there is no dedicated lexicon.[13] In philosophy of language, the view has also been explicitly defended by Rayo (2013), according to whom each expression is associated with a "grab-bag" of different types of information, possibly including "memories, mental images, pieces of encyclopaedic information, pieces of anecdotal information, mental maps and so forth" (2013: 648).

---

[11] I have not mentioned here the addition of *unarticulated constituents*, for example in the apparent addition of a location place when the sentence "it is raining" is uttered. As Recanati (2003: 24–25) notes, it is somewhat unclear to what extent modulation should be distinguished from unarticulated constituent phenomena. For my purposes, I will use "modulation" to refer to all cases of linguistically unlicenced contextual effects on what is said by an utterance, so that unarticulated constituent cases are included. The saturation/modulation distinction can be read then as the licenced/unlicenced context-sensitivity distinction. For a precise definition of linguistic licence, see: (Collins 2020: 1). The distinction also corresponds to the "bottom-up/top-down" distinction employed by Recanati (2003).

[12] An alternative way of using "contextualism" and "radical contextualism" is to interpret the former as restricted to claiming that a certain range of expressions are context-sensitive, while interpreting the latter as claiming that all expressions are context-sensitive. Cappelen and Lepore (2005) arguably take this approach, distinguishing between *moderate* and *radical* contextualism. This has its benefits, as it serves to distinguish the general debate in philosophy of language, from more localized debates in epistemology or ethics, for instance. However, I will follow Recanati's lead in distinguishing the two views as it will prove more useful for our discussion.

[13] See also: (Dilkina et al. 2010).

Finally, before turning back to transformers, it is important to consider two prominent responses to the contextualist position. One kind of response, which Borg (2012) labels *indexicalism*, agrees with the contextualist that context-sensitivity is pervasive, or at least allows that it could be, but denies any form of modulation. Instead, all context-sensitivity is either a form of saturation or is actually some form of pragmatic phenomenon distinct from the truth-conditional content, such as conversational implicature. For example, Stanley (2000) has argued that context-sensitivity is often due to covert variables present in the logical forms of sentences and that these variables can be identified via certain syntactic tests (see: (Collins 2007) for criticism).

The second opposing view is known as *semantic minimalism* (Borg 2004, 2012; Cappelen & Lepore 2005). The minimalist argues that, while the contextualist might be right that what is said in an utterance is often determined partly by contextual processes that are not restricted to saturation, this does not undermine the prospect of a truth-conditional semantics that assigns truth-conditional content to each sentence of the language. The reason for this is that truth-conditional semantics should not capture what is said by an utterance in every case. Instead, it should capture the literal meanings of each sentence by assigning to each a minimal proposition. It may be that the semantic content of a sentence captures what is common across all uses of that sentence (Cappelen & Lepore 2005) or just that it captures what is communicated by that sentence when modulation is completely absent (Borg 2004) or that it captures a certain kind of liability that speakers are subject to in using the words that they did (Borg 2019). A further commitment of minimalism is that the semantic content for each sentence should be recoverable without appeal to speaker intentions. Borg (2012: 11–12), for instance, argues that any appeal to speaker intentions would run counter to the formal ethos that motivates the semantic project in the first place, and furthermore it would be in tension with the plausible idea that semantic competence is realized within a cognitive module (in Fodor's (1983) sense). Semantic minimalism is typically defended on the one hand by emphasizing the need for a truth-conditional semantics, while on the other, showing that the contextualist arguments are in fact ineffective against the view.

With this brief overview of the contextualism debate, we are now in a position to consider the transformer theory i.e. the meaning theory that is implemented in the transformer architecture. The first point to note about the transformer theory is that it is a form of contextualism. The self-attention procedure by design seems to allow for massive amounts of context-sensitivity across occasions of use. An expression meaning as represented by the token representation that is initially assigned to an expression can be modified in whatever way each self-attention head sees fit, given what it has learnt to do through the training procedure. As we saw earlier, no prior restriction is placed on what the self-attention heads are able to do and the self-attention procedure is repeated many times within and across each self-attention layer.[14] A second point to note is that the transformer theory is committed to a form of modulation. There is no particular property or pattern across the token embedding that must be present in order for the self-attention procedure to modify it differently across different contexts. Instead, any expression that passes through the self-attention procedure is effectively guaranteed to be modified in some way. In that respect, the modification that occurs is not linguistically licenced by the token embedding; it is modulation rather than saturation. One initial conclusion we can reach then is that the transformer theory, in allowing for unconstrained context-sensitivity through a process of modulation, rejects indexicalism in favour of contextualism.

[14] Transformer models vary in their number of layers and number of heads within each layer. BERT-base has 12 heads across each of its 12 layers (Devlin et al. 2019). GPT-3 has 96 self-attention heads across each of its 96 layers (Brown et al. 2020).

Given that the transformer theory is contextualist, we should then ask whether it is radically contextualist. Here things become more complicated. It might initially seem plausible that the transformer theory is actually a form of meaning eliminativism. After all, the eliminativist claims that some stored information other than lexical meanings can be employed in interpreting words on a particular occasion, whether that is encyclopaedic knowledge, memory of previous uses, or otherwise. And as we have seen, the transformer approach, as an instance of the more general distributional approach, employs a history of usage for each word as captured in the training corpus. On this basis, it is tempting to think that the transformer theory is a realization of a particular form of meaning eliminativism and thus radical contextualism. After all, on one description of how distributional models work, they make use of the same kind of information that eliminativists like Hintzman (1986) appeal to in their reticence towards word meanings.

Against this thought, however, it is important to keep in mind that, as mentioned earlier, the self-attention procedure is a vector-to-vector system, so if we are to deny that the token embeddings count as appropriate representations of meaning (because they are really just distributional profiles), then we are forced to claim that the contextualized embeddings that it outputs lack any form of meaning as well. But this goes against the assumption made in §1, that transformers do successfully interpret the meanings of sentences and that this achievement is to be located at the contextualized embeddings. Of course, some are skeptical that distributional profiles could serve as representations of meaning, but this is really to be skeptical of the entire distributional semantic approach. As stated earlier, distributional semantics is motivated by the idea that the distributional hypothesis is worth adopting to see what fruit it bears. So the only reason for thinking that the transformer theory is eliminativist is really a reason to reject the entire distributional approach altogether.

It seems, then, that the transformer theory cannot be eliminativist. In fact, focusing on the token embeddings associated with each word reveals an important sense in which the transformer theory is closer to the semantic minimalist position. As we saw earlier, while the radical contextualist claims that expression meanings are not of the right format to figure in the truth-conditional content of a particular occasion of use, the minimalist rejects this claim. But when we turn to what figures as expression meaning and what figures as utterance meaning in the transformer theory, we find that they are similar in important respects. Recall that insofar as we are treating transformers as successfully interpreting sentence meaning, this achievement is located in the contextualized embeddings. The token embeddings that are invariantly assigned to each expression and thus are akin to expression meanings are vectors of the same dimensionality as the contextualized embeddings. So if we are to understand the radical contextualist claim as a technical claim about the type of semantic object associated with an expression, the transformer theory rejects this claim. A possible response is that while the token embeddings might be the same type of object, they might not stand as good representations of a word's meaning in the way that the contextualized embeddings are. That is, it may subsequently be discovered that the token embeddings are in some way deficient or incomplete and that this deficiency is adjusted for in the self-attention process.

Responding to this worry requires investigation of whether the token embeddings used in transformer models possess the same kind of semantic information that one finds in static embeddings such as those produced by Word2vec. While there has been less focus in computational linguistics on the properties of token embeddings compared to contextualized embeddings, there is nevertheless already a wide range of evidence that suggests that token embeddings do encode a good deal of semantic information.

First, token embeddings possess intuitive nearest neighbour relations, where words that are similar in meaning are closer in space, in the same way that static embeddings do. This is illustrated in table 1,[15] where the nearest neighbours in a version of BERT are given for "horse". All expressions in the list are clearly semantically related to "horse" insofar as they pick out similar animal expressions or similar forms of travel:

| **Neighbour** | **Similarity** |
|---|---|
| Horses | 0.68 |
| Dog | 0.43 |
| Cattle | 0.38 |
| Cow | 0.37 |
| Animal | 0.37 |
| Animals | 0.36 |
| Dogs | 0.36 |
| Bike | 0.35 |
| Sheep | 0.35 |
| livestock | 0.35 |

*Table 1: Nearest neighbours of "horse" in BERT-base-uncased*

Focusing on token embeddings in multilingual LLMs, Wen-Yi and Mimno (2023) found that certain multilingual models (specifically the mT5 models developed by Google) will group expressions from a range of languages together, with the very nearest neighbours frequently being translations of one another. Other multilingual models (the XLM-RoBERTa models) separate out expressions from different writing systems but will organize nearest neighbours to be similar in meaning. Turning to a distinct set of findings, Takahashi et al. (2022) found that token embeddings can be used to help predict idiomatic uses of an expression. The intuition behind their approach was that idiomatic uses differ more greatly in meaning from the standard meaning to non-idiomatic uses, and that this would be realized by idiomatic contextualized embeddings being further away from the token embedding than standard contextualized embeddings. Finally, it has also been shown that Word2vec embeddings can actually be used as the token embeddings for a transformer model without loss in accuracy on a document classification task (Suganami & Shinnou 2022). Taken together, these findings suggest that the token embeddings that serve as input to the self-attention procedure can be thought of as meaningful in the same way that contextualized embeddings and Word2vec embeddings are. This runs against the radical contextualist claim that word meanings are in some sense of the wrong format or deficient when compared to utterance meanings.

One significant difference between token embeddings and contextualized embeddings is that the latter contain information about the compositional properties that each word processes. As explained in §2, one thing that the self-attention procedure achieves is to replace a set of token embeddings with a new set of embeddings that captures the way that each word interacts with every other word. That LLMs are able to do this is evident in the findings from Manning et al. (2020) which showed that syntactic trees can be extracted from a transformation of the contextualized embedding space. This difference does not speak to the *wrong format* claim, however, which is concerned with the status of word meanings and not the way that they are combined. That contextualized embeddings also contain compositionality information does not tell us that the way that token embeddings represent word meanings is in any way deficient.

[15] Code for generating nearest neighbours in this paper can be found at https://github.com/Jumblygrindrod/Transf-Cont-Polysemy

This issue of compositionality does reveal a way in which the transformer theory may be quite unlike minimalism. A core part of the minimalist view is a commitment to systematic, compositional semantic processes that take as input the standing meanings of expressions and the syntactic structures for their sentence, and that output some assignment to the sentence. The minimalist is not only committed to invariant word meanings, but invariant sentence meanings as well (putting issues like structural ambiguity to one side). But locating this kind of sentence meaning within the transformer architecture is not a straightforward task partly because locating any notion of sentence meaning is not straightforward. The set of contextualized embeddings for words in a sentence could be taken to represent sentence meaning, except that it will typically contain a great deal more information than just sentential meaning. Specifically, the contextualized embeddings will be sensitive to words outside of the target sentence, such that the contextualized embeddings for the words "Birds" and "fly", will differ across 6 and 7:

6) London is the capital of France. Birds fly.
7) Birds fly.

Manning et al.'s findings that syntactic properties can be extracted from contextualised embeddings are suggestive of the idea that we may come to discover that sentential properties can be extracted from contextualized embeddings, and this may include an invariant notion of sentence meaning. But whether that is the case is an open question. So the transformer theory clearly departs from minimalism in that it doesn't explicitly provide a notion of invariant sentence meaning, but that is partly because the view departs from much of mainstream philosophy of language in not explicitly providing a notion of sentence meaning at all. There likely is some latent notion of sentence meaning encoded across the contextualized embeddings, as is evidenced by the ability of transformer chatbots to identify the meanings of individual sentences, but it remains to be seen what this latent notion looks like.[16]

We can conclude then that the transformer theory makes use of a robust notion of standing meaning assigned to particular expressions that do not seem to be underdetermined in some way. The transformer theory denies the *wrong format* claim that is typical of the radical contextualist view. Instead, the transformer theory seems on this score to be closer to minimalism insofar as minimalism posits just such a notion of standing meaning while allowing that it may not always be present in the communicated content of an utterance. Where the transformer theory departs from minimalism is in the extent of the context-sensitivity it permits. While the transformer theory makes use of token embeddings as a kind of standing meaning, it is near impossible that the token embedding could pass through the self-attention procedure unmodified. In contextualist terms, modulation will effectively always occur. In this respect, the transformer theory manages to reside in a middle ground between minimalism and radical contextualism, rejecting the letter of radical contextualism while maintaining its spirit.

[16] Generating embeddings for sentences through the use of transformer models and similar is a currently an active field. See: Kashyap et al. (2024) for a survey of various approaches.

## 4. Ambiguity within the transformer theory

To further elucidate the transformer theory, I will now turn to a debate closely-related to the contextualist debate in many respects. This concerns how ambiguity should be captured within our account of word meaning. The phenomenon of ambiguity is familiar enough – as the philosopher's favourite example of "bank" will illustrate. But things become more complex as soon as we draw the distinction between polysemy and homonymy. The ambiguity of "bank" is an example of homonymy, where two entirely distinct meanings happen to have the same spelling and pronunciation. On the other hand, polysemy is when a single expression seems to be associated with a group of related senses. For example, "book" can refer to a concrete object that can be thrown, but also to an informational object that cannot be thrown but may well be heart-breaking (e.g. consider the difference between "the book fell on her toe" and "Coetzee's third book was his greatest").

The key question is how exactly polysemy should be treated, and in particular whether it should be treated as distinct in kind from homonymy. The *sense enumeration* approach treats polysemy and homonymy as different only in the number of senses associated with the word. Just as we should approach homonymy by simply listing the various senses that are associated with an expression, we should do the same with polysemy. "Bank" and "book" are alike then in that they both possess at least two lexical entries. But for many, this is an unsatisfactory approach. The fact that no distinction is drawn between homonymy and polysemy on the sense enumeration approach is thought to be problematic given that first, the distinction is intuitive even for the ordinary speaker and second, there are processing differences between polysemous and homonymous expressions found in lexical task tests (Klepousniotou & Baum 2007). Furthermore, many have argued that polysemy is in an important sense *open-ended* – what Recanati (2017: 383) calls the generative aspect of polysemy – in that polysemous senses cannot be counted, as new senses can always be generated on particular occasions of use. It is on this issue of the open-endedness of polysemy that we find a potential overlap with the contextualism debate. Some instances of what contextualists describe as modulation (including 3-5) may potentially be captured as instances of polysemy. Indeed, while there are more specific polysemy phenomena that any good theory needs to capture, including regular polysemy[17] and co-predication,[18] exactly what should be treated as an instance of polysemy is itself a contested issue.

A distinct approach, which arguably has the clearest connection to contextualism, is what Trott and Bergen (2023) label the "core representation" account.[19] On this view, polysemous expressions have a single core meaning, and polysemous readings arise from modifications to the core meaning. Among versions of the core representation account, we can draw a Goldilocks distinction between under-specification, over-specification, and literalist (or *just right*) views. The under-specification view claims that the core meaning is underspecified or deficient in some way, that some further modification must be performed to that meaning in order for it to be associated with a polysemous sense. Obviously this sounds very close to the contextualist picture and this is indeed the picture of polysemy that has previously been advocated by contextualists

[17] Regular polysemy is when a set of polysemous expression have multiple senses that pattern in the same way. For instance, many English animal expressions can be used to refer to the animal or to the animal's meat (e.g. chicken, lamb, rabbit).

[18] Co-predication is when various properties are predicated to a polysemous expression such that multiple polysemous senses appear to be part of the interpretation. For example, in the sentence "the school was built on an ancient church, and had recently been reprimanded by Ofsted", "school" seems to take on both a building sense and an institution sense.

[19] Falkum and Vicente (2015) label this the "one representation hypothesis".

(Recanati 2003: 135). On the other hand, the over-specification view claims that the core meaning is *too rich*, involving too many features or too much information to figure in utterance content. Most notably, Pustejovsky (1998) has argued for a view of this kind, where each polysemous expression is associated with a *qualia structure* and a polysemous reading is generated by selecting just some of the features within the qualia structure. The literalist view claims that the core meaning of a polysemous expression can in fact figure as a polysemous sense, but that equally it can also be modified, giving rise to an indefinite range of polysemous readings.[20] One of the key challenges for the core representation view is being able to outline in a principled way what exactly gets included in the core representation. The over-representation view has previously been accused of including too much information in the lexical entry, and so risks collapse into a kind of meaning eliminativism insofar as there would be no distinction between lexical and non-lexical information (Carston 2012), while the underspecification view and literalist view inherit many of the key points of contention from the contextualism/minimalism debate.

As well as the sense enumeration approach and core representation approach (in its underspecification, overspecification, and literalist forms), the final approach I will consider is the meaning continuity approach (Rodd 2020; Li & Joanisse 2021; Trott & Bergen 2023). According to this approach, meaning is best understood as mapped within a continuous space. Non-ambiguous expressions are represented by a single point in the space. Homonymous expressions will take up two or more distinct points in the space. Polysemous expressions are distinctive insofar as their meaning can be represented across a *region* of the space, where all points in that region are possible polysemic readings.

It might be thought that the meaning continuity view doesn't so much present an alternative theory of ambiguity as provide a distinct modality to represent meaning in (i.e. rather than using discrete symbolic representations, a continuous spatial modality is used). This is only partly right, for even in employing a continuous modality in this way, certain assumptions are built in. For instance, the meaning continuity view is distinct from the sense enumeration approach, for the latter claims that polysemous senses can be captured in a finite list, whereas the former claims that there are an infinite number of distinct uses of a polysemous expression within the expression's region of space. And unlike the core representation view, the meaning continuity view (as I have described it) is not committed to any particular role for a core representation. It may just be that an expression can be used in any of an infinite number of ways that sit within a particular region of that space, without any particular role for a core representation to unite them. Indeed, one significant influence for the meaning continuity approach has been Elman's work on word meaning, according to which word meanings can be modelled as activation patterns within a neural network. The activation pattern for a word depends on the current state of the entire network, and so words will have different meanings on each occasion of use. On such a view, there is no dedicated store of information for each word, and this is emphasised as an advantage for the position insofar as it avoids the challenge faced by the core representation approach of outlining what is included within the core representation (Elman 2009; Trott & Bergen 2023).

One potential drawback of the meaning continuity approach is that it looks ill-equipped to capture the way that uses of a polysemous expression seem to pattern into particular polysemous senses e.g. the object vs informational senses of "book" mentioned earlier. If the meaning continuity view just claims that uses of an expression fall within a particular region in semantic space, no account of these patterns is given. But there are variants of the basic approach that avoid this problem, as Trott & Bergen (2023) detail well. For instance, uses of a polysemous expression could cluster into discrete senses, and transformations of the semantic space could be used to exaggerate those clusters, or those clusters could even be averaged into a single centroid. The focus of this paper though is not on the possibility of capturing such effects, so I will ignore these variants.[21]

[20] Of the core representation views, this position is most congenial to the minimalist view outlined earlier.

It is hopefully immediately apparent that the meaning continuity approach closely resembles the vector-space methodology introduced earlier, and this is not a coincidence. The meaning continuity approach has become more popular in areas such as psycholinguistics partly due to the influence of earlier distributional semantic models such as *Latent Semantic Analysis* (Landauer & Dumais 1997) and , Word2vec.[22] It might be thought then that the transformer theory straightforwardly adopts the meaning continuity approach as well. There is empirical support in favour of this idea. For example, Nair et al. (2020) found a correlation between sense distance in BERT and human judgments of relatedness of senses.[23] So expressions with two ambiguous senses that were judged as unrelated (and thus an instance of homonymy) were also represented in BERT with two centroids quite distant from one another, while for polysemous senses (i.e. two senses of a word that are judged as closely-related), the two centroids were represented in BERT as much closer together.

But treating the transformer theory as simply an instance of the sense continuity position would be too quick. It is certainly the case that in encoding meaning in the form of embeddings, transformer models represent meaning across a continuous modality. However, there is an important sense in which the transformer theory adopts a form of the core representation approach as well. After all, as we have seen, the token embeddings that serve as input to the self-attention process play an important role; they serve as the core representation that is then modified on particular occasions of use. The transformer theory, then, is a combination of two approaches to polysemy that have previously thought to have been distinct.

Given that the transformer theory makes use of a core representation, can we say anything about whether those core representations are over-specified, under-specified, or literalist? I take it this would be an interesting empirical topic to explore, but currently I know of no empirical work that obviously sheds light on this question. We would want to know whether the self-attention procedure effectively removes, adds or modifies the information contained within the token embeddings. One possibility worth considering is that the distinction between over-specification and under-specification is not easily employed when considering the relationship between token embeddings and contextualized embeddings, for the identity conditions for particular pieces of information start to become somewhat unclear. But I will not pursue that line of thought here, leaving this issue as an open question.[24]

It is worth noting a particularly surprising feature of this approach: it draws no strong distinction between homonymy and polysemy in terms of how each is realized across token and contextualized embeddings. The homonymy of expressions like "bank" and "rock" are treated in

[21] Trott & Bergen show that there are accuracy and reaction time effects on tasks with polysemous expressions that are better explained by such variants (what they call "hybrid views").
[22] See: (Günther et al. 2019) for a useful overview.
[23] For the purposes of their study, senses were generated by taking the mean of the set of vectors that all belong to the same polysemous sense. An average vector of this kind is called a *centroid*.
[24] Thanks to an anonymous reviewer for raising this issue.

the same way that polysemy is treated: the token embedding serves as a core representation that is modified through the self-attention procedure to generate the different readings. The token embedding will somehow have to do justice to each of its homonymous senses. This is illustrated quite strikingly by looking at the nearest neighbour list of the token embedding for a homonymous expression like "rock", which indicates that somehow the token embedding resides halfway between geological terms and musical terms:

| Neighbour | Similarity |
|---|---|
| Rocks | 0.6 |
| Stone | 0.46 |
| Metal | 0.38 |
| Pop | 0.37 |
| Rocky | 0.36 |
| Stones | 0.36 |
| Punk | 0.34 |
| Boulder | 0.34 |
| Cliff | 0.34 |
| Wood | 0.32 |

*Table 2: Nearest neighbours of "rock" in BERT-base-uncased*

So while the token embeddings seem to play a role akin to the intuitive notion of standing or literal meaning of the expression-type, how this works in the case of homonymous expressions seems quite counterintuitive. The criticism that is usually levelled against the sense enumeration approach – that polysemy is not treated as distinct from homonymy – can actually be levelled against the transformer theory here. But while the sense enumeration approach seems guilty of applying a plausible approach for homonymy to polysemy, the transformer theory seems guilty in the other direction – of applying a plausible approach for polysemy to homonymy. All that said, we don't really know much yet about how or what is being encoded into these token embeddings, and so we cannot rule out that the model has found a way to encode within a particular vector *that it is n-ways homonymous*. Exploring how ambiguity can be handled within a continuous vector space is currently a flourishing area of research, and the findings of (Yaghoobzadeh et al. 2019) are suggestive here. They investigated whether a probing classifier can be trained that predicts the sense class of a particular expression.[25] They also investigated whether a classifier can be trained to predict whether an expression is ambiguous at all – where ambiguity is operationalized as belonging to more than one sense class. They were able to produce a highly accurate classifier on this latter task. Note something quite particular about this study: there has been previous work that has sought to show that ambiguity in embeddings can be predicted on the basis of an embedding's geometric relationship to other embeddings (Wiedemann et al. 2019). But Yaghoobzadeh et al.'s approach probes single embeddings directly, rather than relationships between vectors, and so this is stronger evidence that the ambiguity of the expression has been encoded directly into the embedding. However, their study used static embeddings (like Word2vec), so it remains to be seen whether the same applies to the token embeddings that are used by transformer models.

[25] A sense class is a kind of general category with which we can distinguish between different senses of a word. For instance, included in Yaghoobzadeh et al.'s sense classes are ORGANIZATION and FOOD. With these two categories, we can distinguish between two different senses of "apple". Yaghoobzadeh et al. used 34 sense categories determined using Wikipedia links.

## 5. Conclusion

I want to conclude by resituating the findings of this paper within the larger project outlined in §1. We have treated the transformer architecture here as if it is successful in interpreting linguistic meanings by assigning contextualized embeddings to words, and we have then explored the commitments of the meaning theory that is implemented within the architecture. The transformer theory combines the insights of both minimalism and contextualism in its general approach to context-sensitivity by allowing for a notion of standing meaning as playing an important role in content generation while also allowing for contextual processes so extensive that the standing meaning will effectively never figure in utterance content. Regarding polysemy, we again find that the insights of two popular approaches are combined: like the core representation approach, polysemous readings are generated by modifying a meaning that is associated with an expression; like the meaning continuity approach, polysemous readings are open ended insofar as a particular expression will allow for an infinite number of polysemous readings that reside within a region of semantic space.

Alongside these specific results regarding the workings of the transformer architecture, I hope this paper has shown that investigation of the way in which artificial systems process linguistic data can be of value to linguistic theorizing without being committed to the idea that such systems must reflect human linguistic cognition. Exploring such systems allows us to consider specific issues (on polysemy and context-sensitivity, and more much beside) through the construction of test cases and the evaluation of possibility claims. It also allows us to consider ways of processing linguistic data that at first glance appear unintuitive or implausible but that may be surprisingly successful (as we have seen with the transformer theory's treatment of homonymy). I do think that the remarkable advances we have seen in natural language processing mean that further inquiry into the plausibility of the transformer theory is warranted. But even if the transformer theory is not, in the end, plausible, the wider project of bringing insights from language model technology to bear against philosophical and linguistic theorizing extremely worthwhile and more achievable than ever.

**Acknowledgements**: I would like to thank Nat Hansen, Emma Borg, Sean Trott, and audiences at the Philosophy of AI conference in Erlangen and at the High Performance Computing Centre (HLRS) Stuttgart. I would also like to thank the editors and referees at *Ergo*, who provided very useful comments and as a result improved the paper a great deal.